\documentclass[letterpaper, 10 pt, conference]{ieeeconf}  

\IEEEoverridecommandlockouts                              

\usepackage[dvipsnames]{xcolor}

\usepackage{lineno,hyperref}

\usepackage{graphicx}

\usepackage{soul}

\setstcolor{Green}

\title{\LARGE \bf
Weakly-Supervised Object Detection Learning through\\ Human-Robot Interaction
}

\author{Elisa Maiettini$^{1}$ \and Vadim Tikhanoff$^{2}$ \and Lorenzo Natale$^{1}$
	\thanks{$^{1}$ Humanoid Sensing and Perception, Istituto Italiano di Tecnologia, Genoa, Italy}%
    \thanks{$^{2}$ iCub Tech, Istituto Italiano di Tecnologia, Genoa, Italy}
}

\begin{document}

\maketitle
\thispagestyle{empty}
\pagestyle{empty}

\begin{abstract}

Reliable perception and efficient adaptation to novel conditions are priority skills for humanoids that function in dynamic environments. The vast advancements in latest computer vision research, brought by deep learning methods, are appealing for the robotics community. However, their adoption in applied domains is not straightforward since adapting them to new tasks is strongly demanding in terms of annotated data and optimization time. Nevertheless, robotic platforms, and especially humanoids, present opportunities (such as additional sensors and the chance to explore the environment) that can be exploited to overcome these issues.

In this paper, we present a pipeline for efficiently training an object detection system on a humanoid robot. The proposed system allows to iteratively adapt an object detection model to novel scenarios, by exploiting: (i) a teacher-learner pipeline, (ii) weakly supervised learning techniques to reduce the human labeling effort and (iii) an on-line learning approach for fast model re-training. We use the R1 humanoid robot for both testing the proposed pipeline in a real-time application and acquire sequences of images to benchmark the method.
We made the code of the application publicly available.

\end{abstract}


\section{INTRODUCTION}
\label{sec:intro}


Much research on robot vision draws from techniques developed in computer vision. However, robot applications have specific requirements.
Particularly, high reliability and fast adaptation are both fundamental requisites of vision systems for robots that need to operate in unconstrained and dynamic environments. 
Mainstream computer vision solutions typically rely on deep learning based models which have large amount of parameters that need to be tuned at training time. This has well known implications: firstly, the amount of effort required to provide annotations for the, typically, very large training set and, secondly, long training time. The remarkable performance achieved with such techniques are, therefore, appealing for robotics, but their adoption in robotic applications remains limited to those problems for which annotated, large datasets are available and there is no need for on-line adaptation or re-training.

Nevertheless, robotics offers some opportunities that are often unexplored in the literature. For instance, the embodiment of the robot can be exploited to interact with the environment, including humans, to actively acquire training data. Moreover, since robots are frequently equipped with multiple sensory modalities, there is additional information that can be used to aid learning. This is especially true for humanoid robots, such us the iCub~\cite{icub} and R1~\cite{r1}, which are equipped with depth, force or tactile sensors.

In this paper, we describe the implementation of a complete pipeline for training an object detection system, embedded on a humanoid robot. The system allows to train and iteratively adapt an object detection model, with minimal labeling effort. To do so, the system is designed to conveniently exploit (i) the interaction with a human teacher, (ii) a fixed set of exploratory behaviors and (iii) a weakly-supervised learning algorithm, to reduce the number of frames that are manually annotated while preserving performance. To achieve a smooth interaction between the robot, the environment and the teacher, the system also integrates an architecture for object detection that can be quickly re-trained on-line. 
We implemented the system on the R1 humanoid robot~\cite{r1}, to demonstrate its performance in a real-time application. The functioning of the system is demonstrated in the video provided as supplementary material. In addition, we extensively benchmark the system using a series of sequences that were recorded and manually annotated to provide ground-truth, demonstrating the effectiveness of the pipeline in reducing the annotation effort, while retaining performance. Finally, we made the code of the application publicly available\footnote{\url{https://github.com/robotology/online-detection-demo}} for reproducibility.

The remaining of this paper is organized as follows: in Sec.~\ref{sec:relwork}, we introduce the state-of-the-art of current object detection in robotics. In Sec.~\ref{sec:methods}, we present the components of the proposed pipeline, which are analyzed and validated in Sec.~\ref{sec:experiments}. Finally, Sec.~\ref{sec:conclusions} concludes the paper.

\begin{figure}[]
	\centering
	\includegraphics[width=0.45\textwidth]{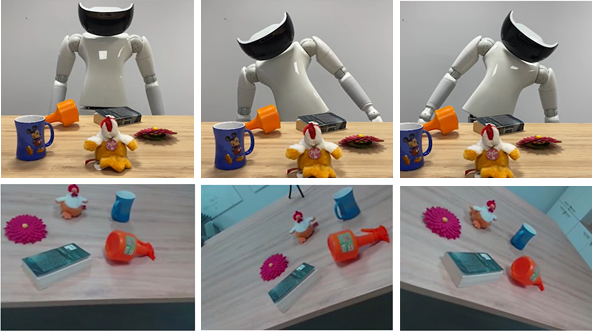}
	\caption{Example frames of the R1's exploratory behaviors used in the proposed application. Different R1's poses (first row) allows to acquire different views of the objects of interest (second row).}
	\label{fig:exploration}
\end{figure}


\section{RELATED WORK}
\label{sec:relwork}

Latest research on object detection for robotics mainly focuses on improving precision in particularly difficult scenarios such as, occlusion and clutter~\cite{Zeng2018,Georgakis2017,Schwarz2018,Tobin2017,Chen2020}. To this aim, a widely used approach is to rely on deep learning based methods (like e.g. Faster R-CNN~\cite{ren2015_faster} and its evolution, Mask R-CNN~\cite{He2017}, YOLO~\cite{yolact}, Centernet~\cite{duan2019} and Cornernet~\cite{law2018}) which demonstrated remarkable performance on general purpose~\cite{pascal2010,coco,imagenet} and robotics~\cite{Calli2015} datasets. A common approach in the state-of-the-art, characterizing the so-called \textit{region-based} approaches, is that of splitting the detection pipeline into three main stages: (i) generation of region candidates that might contain the objects of interest, (ii) region proposals encoding into convolutional features and (iii) region classification and refinement. In the last years, the common trend has been to integrate these three stages into “monolithic” models, trained end-to-end via back-propagation~\cite{ren2015_faster,He2017,Shrivastava2016,dai2016}. A drawback of such architectures is 
that the training process requires large annotated datasets and long optimization time, and it is, therefore, badly suited for robots operating in dynamic environments. In the following paragraphs, we cover the work that has been done in the literature in order to relieve deep learning based methods requirements in terms of training time and annotated data.\\

\begin{figure*}[]
	\centering
	\vspace{3mm}
	\includegraphics[width=0.9\textwidth]{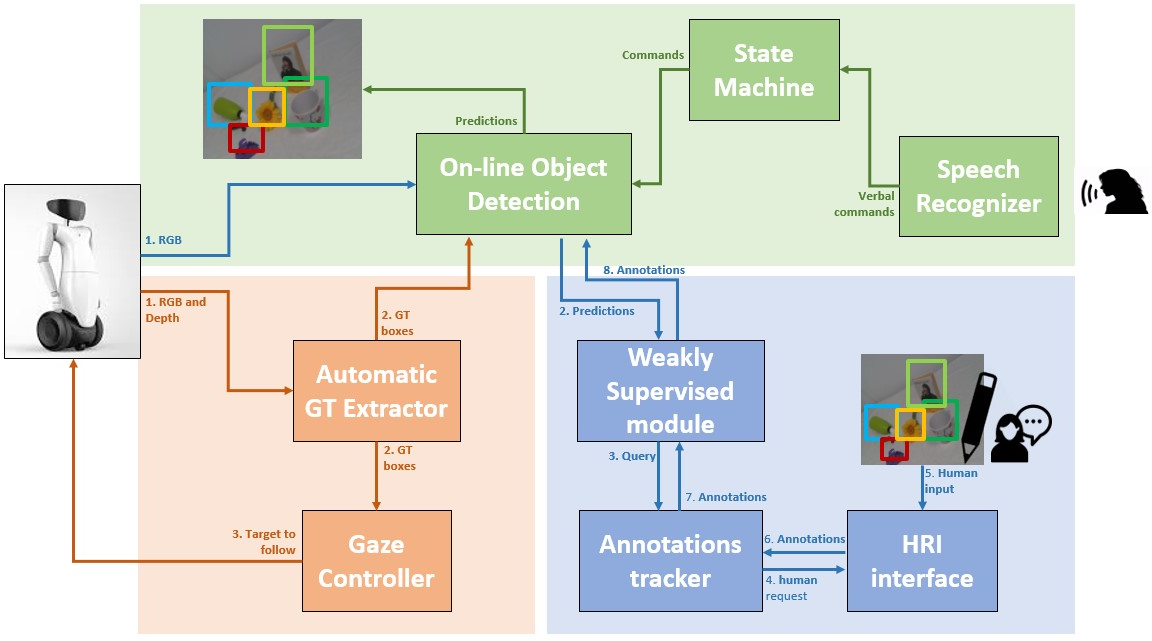}
	\caption{Architecture of the proposed pipeline. The green block represents the \textit{Object detection system}, the red block is the \textit{Automatic in-hand supervision extraction} and finally, the blue block depicts the \textit{Weak-supervision system}. The flow of red arrows represents the \textit{Supervised training phase}, the blue one represents the \textit{Refinement phase} and the green one is in common between the two phases. Refer to Sec.~\ref{sec:methods} for further details.}
	\label{fig:architecture}
\end{figure*}

\noindent{{\bf Time-efficient object detection learning. }}
Even though the major trend is to design detection models as  monolithic architectures, trained end-to-end, another natural approach is to consider multi-stage architectures, tackling each step of the detection pipeline separately (see e.g., ~\cite{girshick2014_rcnn,girshick15_fastrcnn} and specifically~\cite{maiettini2018, maiettini2019a}). As showed in~\cite{maiettini2019a}, this latter approach is key to obtain fast adaptation capabilities to a novel task. Specifically, the proposed on-line method~\cite{maiettini2019a} is composed of a deep learning-based region proposal and feature extractor (based on Faster R-CNN~\cite{ren2015_faster}), followed by a Kernel based method for classification and a boostrapping approach to address the well known issue of background-foreground imbalance in object detection~\cite{Shrivastava2016}. This pipeline is suited for a typical unconstrained robotic setting, as, while keeping the pre-trained feature extractor fixed, the combination of an efficient classifier (FALKON~\cite{falkon2018,falkonlibrary2020}) with an approximated hard negatives bootstrapping (see~\cite{maiettini2019a}) provides fast model training. Moreover, lately, the same approach has been adapted to the region proposal generation~\cite{ceola2020a} and object instance segmentation~\cite{ceola2020b}. In this work, we build on~\cite{maiettini2019a}, proposing a unified pipeline to rapidly train and update a detection model on novel tasks and scenarios, with different human robot interactions for ground-truth acquisition.\\

\noindent{{\bf Data-efficient object detection learning. }}
In past work~\cite{maiettini2017}, the annotation problem has been addressed by exploiting a human robot interaction to collect automatically annotated images. Specifically, within a teacher-learner pipeline~\cite{pasquale2016_frontiers}, motion and depth cues are used to follow with the robot's gaze the object, segment it and automatically assign the correspondent bounding box. While allowing for a natural interaction and accurate object detection~\cite{maiettini2017}, this pipeline limits the generalization capabilities to novel, unseen, scenarios~\cite{maiettini2020}. This issue has been addressed with promising results in~\cite{maiettini2019b,maiettini2020} by exploiting (i) the large amount of un-labeled images acquired by the robot during operation and (ii)  the weakly-supervised learning family of techniques (WSL)~\cite{Zhou2017,Hernandez2016a}. In WSL, the most relevant sub-classes of methods, for this work, are \textit{Active Learning} (AL) and \textit{Self-supervised Learning} (SSL). In AL~\cite{settles2012}, the effort is focused on defining a sample selection policy to choose the most informative samples to be asked for annotation to a human. Recently, AL has been successfully applied to the object detection task~\cite{Aghdam_2019,haussmann2020,desai2019,kao2018}. SSL, instead, attempts to exploit the unlabeled instances without querying human experts, by e.g., using the high-confident predictions as pseudo ground-truth~\cite{huang2020}. This latter family of approaches does not imply any human annotation, however it may suffer of model drift and degradation due to errors in the self-supervision. Conversely, AL, while being more stable, still requires a human labeling. Recent approaches integrate both AL and SSL into the same detection pipeline, like e.g., the Self-supervised Sample Mining (SSM)~\cite{Wang2018,wang2019}. Finally, in~\cite{maiettini2020}, the SSM has been integrated with the aforementioned fast object detection learning~\cite{maiettini2019a}. While the approach is promising, the required number of manually labeled images (in the order of a few hundreds for a task of 30 objects~\cite{maiettini2020}), prevents from performing the model refinement on-line. In this work, we improve this aspect by integrating a bounding box tracker~\cite{farhadi2018}, to further reduce the number of annotations required, by propagating the provided labels between consecutive frames. Moreover, we present a unified system for on-line training and updating an object detection model, which allows to adapt to novel tasks and scenarios with active exploration and human robot interaction.


\section{METHODS}
\label{sec:methods}


In this work, a robot is asked to detect a set of object instances in an unconstrained environment. The aim of the presented application is that of providing a system that allows training and subsequently updating an object detection model. This is done by means of different human robot interaction modalities, while reducing annotation effort by leveraging on autonomous explorative behaviors and a weakly-supervised learning strategy.

\subsection{Overview of the pipeline}
The proposed pipeline is initially trained during a brief interaction with a human, in a constrained scenario (hereinafter referred to as \textit{Supervised training phase}). The goal of this initial interaction is that of ``bootstrapping'' the system, with an initial object detection model. After that, the robot relies on this initial knowledge to adapt to new settings, by actively exploring the environment and asking for limited human intervention (hereinafter referred to as \textit{Refinement phase}). The \textit{Refinement phase} iteratively builds new training sets. Using the initial detector, the robot detects candidate bounding boxes on the incoming images. These predictions are evaluated to determine their confidence. High confidence labels are added to the training set, with a SSL strategy. Instead, when low confidence bounding boxes are detected, the robot stops the exploration and asks the human teacher to either accept or refine them, with an AL strategy. The teacher provides refined annotations using a graphical interface on a tablet. The new annotations are used to initialize a set of trackers, which propagate bounding boxes as the exploration resumes.

To summarize, the proposed application, is represented in Fig.~\ref{fig:architecture}. This can be divided in three main components: (i) the \textit{Object detection system} (green blocks), (ii) the \textit{Automatic in-hand supervision extraction} (red blocks) and (iii) the \textit{Weak-supervision system} (blue blocks). We rely on the YARP\footnote{\url{https://www.yarp.it}} middleware for connecting the different modules. In this section, we describe the main components of the presented pipeline.

\subsection{Object detection system}
\label{methods:ood}
The \textit{Object detection system} is composed of three main modules: (i) the \textit{State Machine}, (ii) the \textit{Speech recognizer} and (iii) the \textit{On-line object detection}. The former one, which is implemented in LUA\footnote{\url{https://www.lua.org/}} and rFSM\footnote{\url{https://github.com/kmarkus/rFSM}}, orchestrates the different components of the application. There are three main states: (i) \textit{Inference}, which is the default state, (ii) \textit{Supervised train} and (iii) \textit{Weakly-supervised train}. The different states are triggered by the user's verbal commands or questions which are received by means of the \textit{Speech recognizer} module. The \textit{On-line object detection} is described in details in the following paragraphs. \\

\noindent{{\bf On-line object detection.}}
This module detects the objects of interest from the stream of images from the robot's cameras. For its implementation, we follow the on-line object detection learning method proposed in~\cite{maiettini2019a}. This is a \textit{region-based} approach (see Sec.~\ref{sec:relwork}) which consists of two stages: (i) region proposals and feature extraction, and (ii) region classification and bounding box refinement. The first stage relies on layers from Faster R-CNN~\cite{ren2015_faster} (specifically, the convolutional layers, the Region Proposal Network~\cite{ren2015_faster} and the \textit{RoI Pooling layer}~\cite{girshick15_fastrcnn}). This part is used to extract from an image a set of candidate regions that might contain the objects of interest, so-called Regions of Interest (RoIs), and encode them into a set of convolutional features. The second stage is composed of a set of Kernel based binary classifiers (one for each class of the considered detection task) and Regularized Least Squares (RLS)~\cite{hastie_elements_2009}, respectively for the classification and the refinement of the proposed RoIs. We used the recently proposed FALKON~\cite{falkon2018,falkonlibrary2020} for region classification.
While the first stage is trained only once and off-line on the available data, the second one can be updated on-line as new data come. Specifically, the classifiers are trained with an approximate bootstrapping approach, called Minibootstrap~\cite{maiettini2019a}, which addresses the well-known issue in object detection of background-foreground class imbalance, while maintaining a short training time.

During the \textit{Supervised training phase} (represented by the flow of red arrows in Fig.~\ref{fig:architecture}), the application is in the \textit{Supervised train} state and the detection system processes the incoming images and labels, training a new FALKON classifier and a new RLS refiner for each novel class. During the \textit{Refinement phase} (represented by the flow of blue arrows in Fig.~\ref{fig:architecture}), instead, the application's state is \textit{Weakly-supervised train} and the detection model is refined on the new scenario. Firstly, the current detection model predicts the incoming unlabeled images. Then the \textit{Weakly-supervised module} evaluates these predictions (as described in Sec.~\ref{methods:ws}), adding the confident ones to the dataset as self-supervised pseudo ground-truth (with a SSL strategy) and asking for manual annotation for the uncertain ones (with an AL strategy). When the acquisition is completed, the \textit{On-line object detection} updates the FALKON classifiers and the RLS models on the novel data.
Finally, during the \textit{Inference} state, the detection system receives the stream of images as input and provides a list of predicted detections as output.

\subsection{Automatic in-hand supervision extraction}
\label{methods:supervised}
This component (red block in Fig.~\ref{fig:architecture}) implements the procedure for the automatic annotation acquisition for handheld objects which is used during the \textit{Supervised training phase}. It is composed of the \textit{Automatic ground-truth extractor}\footnote{Module taken from \url{https://github.com/robotology}\label{robotology}} and the \textit{Gaze controller}$^{\ref{robotology}}$ (see Fig.~\ref{fig:architecture}). For the former we rely on~\cite{maiettini2017}. This pipeline exploits the depth information and human-robot interaction in order to segment the blob of pixels belonging to the object of interest. Specifically, the human shows the object in front of the camera of the robot. A tracking routine~\cite{pasquale2016_frontiers} selects the pixels from the depth map that are closer to the robot, segmenting them from the background. It computes a bounding box surrounding this particular blob and sends it to the \textit{Object detection system} and to the \textit{Gaze controller}. The former uses the bounding box as ground-truth for training a new detection model during the \textit{Supervised training phase}, while the \textit{Gaze controller} uses this information to generate a target point to follow in order to make the robot track the object of interest. 

\subsection{Weak-supervision system}
\label{methods:ws}
This component (blue block in Fig.~\ref{fig:architecture}) implements the weakly-supervised learning strategy that allows for the detection model's refinement during the \textit{Refinement phase}. This is composed by four main modules: (i) the \textit{Exploration module}, (ii) the \textit{Weakly-supervised module}, (iii) the \textit{Annotations tracker} and (iv) the \textit{Human-Robot Interaction (HRI) interface}.\\

\noindent{{\bf Exploration module.}} This module makes the robot follow a specified trajectory, with the aim of exploring the surrounding environment, acquiring different views of the objects. This exploration needs to be paused when the human intervention is required to provide manual annotation and resumed when the labeling process is accomplished. For this work, we considered a fixed set of exploratory movements for the upper body of the robot to acquire new views of the objects. Refer to Fig.~\ref{fig:exploration} for examples frames of the R1's exploratory movements used for the proposed pipeline. More sophisticated actions may be performed, for example, to actively manipulate objects.\\

\noindent{{\bf Weakly-supervised module. }} During the exploration, this module receives the predictions of the stream of images from the \textit{On-line object detection}, it evaluates them and decides whether to use them as self-supervision (with a SSL strategy) or to ask for an external labeling (with an AL strategy). For this module, we relied on the weakly-supervised strategy proposed in~\cite{maiettini2019b} and developed in~\cite{maiettini2020}. Specifically, first a \textit{Scoring function} assigns a confidence score to each unlabeled image, based on the received predicted detections. This confidence score is, then, used by a \textit{Selection policy} to decide whether an image needs to be queried for annotation or the predicted detections are confident enough to be used for self-supervision. Note that, we simplify the \textit{Scoring function} with respect to the one used in~\cite{maiettini2020} to reduce the per-image processing time. Specifically, instead of using the so-called \textit{Cross-image validation} (see~\cite{Wang2018} and~\cite{maiettini2020}), the confidence score of the image is obtained by averaging the confidence scores of the single predictions. Then, similarly to previous work, the \textit{Selection policy} compares the per-image confidence scores with two thresholds, namely $th_l$ and $th_h$. If the confidence score is lower than the $th_l$, the image is considered doubtful and thus asked for annotation (AL). If instead the score is higher than $th_h$ the predictions in the image are considered confident and thus can be used as pseudo ground-truth for refinement (SSL). All the other predicted images are not used. Using an average score for the whole image in some cases may lead the system to use bounding boxes whose individual score is low. To avoid this situation, we also introduce a minimum threshold (namely, $th_m$) and verify that all bounding boxes within an image have a score that is at least above this minimum threshold, if this is not the case, the whole image is marked for annotation regardless the average confidence score.\\
\begin{figure*}[]
	\centering
	\vspace{3mm}
	\includegraphics[width=0.9\textwidth]{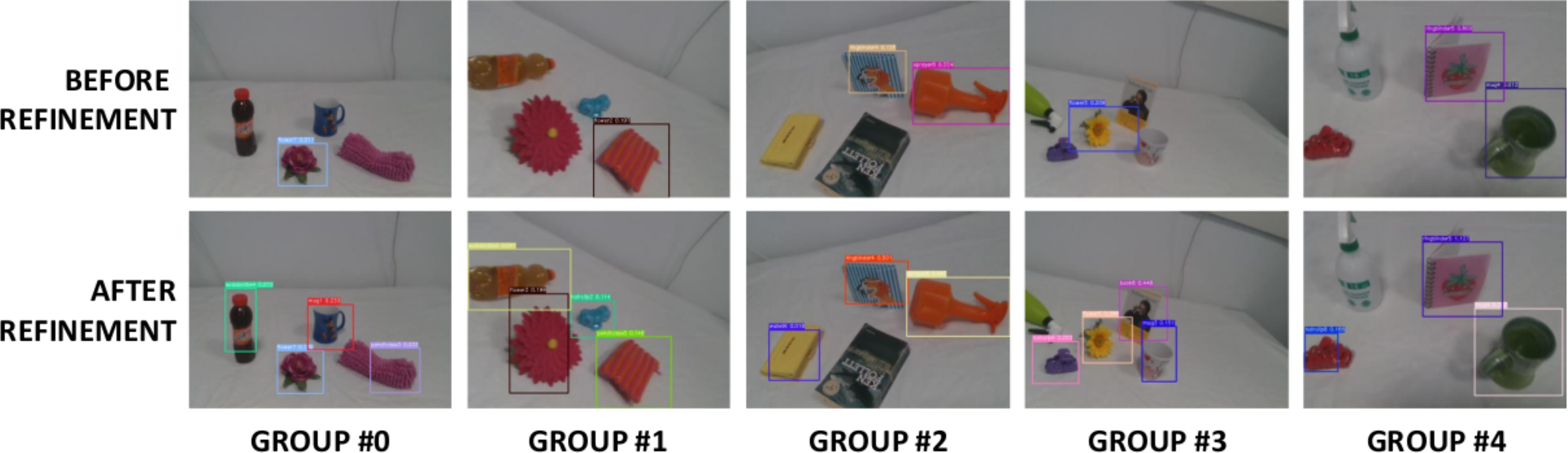}
	\caption{Examples of detected images, from the 5 objects groups, obtained by using the \textsc{Before refinement model} and the \textsc{After refinement model} on new unseen sequences. See Sec.~\ref{experiments:generalization} for details.}
	\label{fig:detections}
\end{figure*}

\noindent{{\bf HRI interface and Annotations tracker. }}Finally, one of the main contributions of this work relies on the \textit{HRI interface} and \textit{Annotations tracker} modules. The former simplifies the annotation process, while the latter propagates the labels provided by the human across consecutive frames. Specifically, when the \textit{Weakly-supervised module} requires annotations for an image,
the \textit{HRI interface} is activated, so that the teacher can annotate the frame. 
This is done with a smooth labeling procedure: the \textit{HRI interface} receives verbal commands and it opens an interactive window where the user can draw the correct bounding boxes or refine the ones predicted by the detection model. Finally, provided annotations are used to initialize the \textit{Annotations tracker} which propagate them in future frames, further reducing the human labeling effort. For this module, we rely on the multi-object tracker \textit{Re3}~\cite{farhadi2018}, which we integrated in our application. To avoid the tracker to diverge in case of extreme clutter or occlusions, causing a deterioration of the bounding boxes, we evaluate their quality by computing the overlap between different objects. When the overlap is too high, the quality is considered low. In this case, the robot asks the human teacher to provide a new set of annotations and the tracker is re-initialized. In Sec.~\ref{experiments:ws_analysis}, we quantify the precision of the bounding boxes provided by the tracker and demonstrate that this allows to remarkably reduce the manual annotation effort.\\


\section{EXPERIMENTS}
\label{sec:experiments}

While the video submitted as supplementary material shows the functioning of the proposed application, in this section, we demonstrate its effectiveness reporting on the performed empirical evaluation. The experimental validation focuses on the refinement pipeline, because the validation of the initial \textit{Supervised training phase} has been explored in previous work~\cite{maiettini2017,maiettini2019a}. We consider, therefore, the refinement following the initial supervised training of the detector by means of the \textit{Automatic in-hand supervision extraction}, presented in Sec.~\ref{methods:supervised}. This model needs to be refined in order to generalize to a different setting (i.e., a table top) by exploiting unlabeled data collected by the robot during exploration (by means of the \textit{Weak-supervision system}, presented in Sec.~\ref{methods:ws}).

Given the focus of the experimental validation, for the \textit{Supervised training phase}, we use previously acquired data, i.e. the iCubWorld-Transformations dataset~\cite{pasquale2019} (iCWT). iCWT contains images of 200 objects, demonstrated by a human teacher to the robot, within the conditions described in Sec.~\ref{methods:supervised}. To benchmark the performance of the system after the \textit{Refinement phase} and compare the detection performance, we selected 21 objects, that are part of the 200 objects in iCWT, and divided them in 5 groups. For each group, the objects were placed on a table in two different arrangements. Then, each set was presented to the robot for the \textit{Refinement phase}. During the exploration, the sequence of images were recorded and, subsequently, manually annotated. Manual annotation of the sequences was required to obtain ground-truth for benchmarking purposes, i.e. the ground-truth acquired in this way was not used for training (we call these sequences \textsc{table-top} in the remainder of the paper).
\begin{table}[t]
	\centering
	\caption{This table reports on the refinement capabilities of the proposed approach, comparing \textsc{Before refinement model} with the \textsc{After refinement model}. Refer to Sec.~\ref{experiments:evaluation} for details.}
	\begin{tabular}{|c|c|c|c|}
		\cline{1-4}
		\multicolumn{1}{|c|}{\textbf{\begin{tabular}[c]{@{}c@{}}\textbf{Objects}\\ \textbf{group}\end{tabular}}} & \multicolumn{1}{c|}{\textbf{\begin{tabular}[c]{@{}c@{}}Before ref.\\ (mAP(\%))\end{tabular}}} & \multicolumn{1}{c|}{\textbf{\begin{tabular}[c]{@{}c@{}}After ref.\\ (mAP(\%))\end{tabular}}} & \multicolumn{1}{c|}{\textbf{\begin{tabular}[c]{@{}c@{}}Manual\\ annotations\end{tabular}}} \\ \hline
		\multicolumn{1}{|c|}{\textbf{\#0}} &         14.4\%              &        90.6\%              &      4 images (12 bbox)\\ \hline
		\multicolumn{1}{|c|}{\textbf{\#1}} &         58.2\%              &        92.9\%              &      3 images (10 bbox)\\ \hline
		\multicolumn{1}{|c|}{\textbf{\#2}} &         44.1\%              &        95.1\%              &      4 images (12 bbox)\\ \hline
		\multicolumn{1}{|c|}{\textbf{\#3}} &         36.2\%              &        78.4\%              &      4 images (12 bbox)\\ \hline
		\multicolumn{1}{|c|}{\textbf{\#4}} &         48.5\%              &        87.9\%              &      5 images (13 bbox)\\ \hline
	\end{tabular}
	\label{table:evaluation}
\end{table}
In the following sections, we report results for 5 different experiments, one for each objects group in the \textsc{table-top}. 
During each experiment, the robot performed a refinment following the \textit{Refinement phase} (see Sec.~\ref{sec:methods}). Note that, during this latter phase a human manually provides the annotations, by means of the \textit{HRI interface}. 

In the reported experimental analysis, for the \textit{On-line object detection}, we adopt the CNN model proposed in~\cite{matthew2013_zf} as convolutional backbone for the Faster R-CNN based feature extractor. This has been trained on a 100 objects identification task from the iCWT (as explained in~\cite{maiettini2019a}) and we used the learned weights for the subsequent on-line learning steps. Note that, the 100 objects have been chosen excluding the 21 in the \textsc{table-top}. We rely on~\cite{maiettini2020} to set the hyper-parameters of FALKON and the Minibootstrap. Then, for the \textit{Weakly-supervised module}, we empirically set the $th_l=0.3$, $th_h=0.4$ and $th_m=0.1$.

We report performance in terms of mAP (mean Average Precision) with the IoU (Intersection over Union) threshold set to 0.5, as defined for Pascal VOC 2007 (see~\cite{pascal2010} for further details).

\begin{table*}[t]
	\centering
	\vspace{3mm}
	\caption{This table analyzes the impact of the different components of the Weak-supervision system. Refer to Sec.~\ref{experiments:ws_analysis} for details.}
	\begin{tabular}{|c|c|c|c|c|c|}
		\cline{1-6}
		\multicolumn{1}{|c|}{\textbf{\begin{tabular}[c]{@{}c@{}}\textbf{Objects}\\ \textbf{group}\end{tabular}}} & \multicolumn{1}{c|}{\textbf{\begin{tabular}[c]{@{}c@{}}Manual\\ annotations\end{tabular}}} & \multicolumn{1}{c|}{\textbf{\begin{tabular}[c]{@{}c@{}}Total AL\\ queries\end{tabular}}} & \multicolumn{1}{c|}{\textbf{Total SSL}} & \multicolumn{1}{c|}{\textbf{\begin{tabular}[c]{@{}c@{}}Tracker\\ (mAP(\%))\end{tabular}}} & \multicolumn{1}{c|}{\textbf{\begin{tabular}[c]{@{}c@{}}Pseudo labels\\ (mAP(\%))\end{tabular}}} \\ \hline
		\multicolumn{1}{|c|}{\textbf{\#0}} &        4 images (12 bbox)       &       200 images (800 bbox)    &  0     & 88.0\% & 88.0\%  \\ \hline
		\multicolumn{1}{|c|}{\textbf{\#1}} &        3 images (10 bbox)       &       173 images (692 bbox)    &  5     & 97.4\% & 92.9\% \\ \hline
		\multicolumn{1}{|c|}{\textbf{\#2}} &        4 images (12 bbox)       &       197 images (788 bbox)    &  0     & 97.0\% & 97.0\% \\ \hline
		\multicolumn{1}{|c|}{\textbf{\#3}} &        4 images (12 bbox)       &       134 images (670 bbox)    &  24    & 84.4\% & 76.0\% \\ \hline
		\multicolumn{1}{|c|}{\textbf{\#4}} &        5 images (13 bbox)       &       163 images (652 bbox)    &  1     & 95.3\% & 95.4\% \\ \hline
	\end{tabular}
	\label{table:ws}
\end{table*}

\subsection{Detection refinement evaluation}
\label{experiments:evaluation}
For each objects group previously described, we define the two following detection models:
\begin{itemize}
	\item \textsc{Before refinement model}, which is obtained after the \textit{Supervised training phase} as described previously.
	\item \textsc{After refinement model}, which, instead, is obtained after the \textit{Refinement phase}, updating the detection model, following each \textsc{table-top} exploration sequences.
\end{itemize}
The obtained results are reported in Tab.~\ref{table:evaluation}. For each of the 5 objects group, we report the accuracy on one of the \textsc{table-top} sequences, obtained by (i) the \textsc{Before refinement model} (\textbf{Before ref.} column in Tab.~\ref{table:evaluation}) and (ii) the \textsc{After refinement model} which has been refined during that exploration sequence (\textbf{After ref.} column in Tab.~\ref{table:evaluation}). Finally, we report the number of manual annotations required to the human, both in terms of images and manually drawn bounding boxes (\textbf{Manual annotation} column in Tab.~\ref{table:evaluation}).

As it can be observed, for all the 5 objects groups, as expected, the \textsc{Before refinement model} performs poorly on the new table top scenario. The reason for this is twofold. Firstly, there is a considerable domain shift between the \textit{Supervised training phase} (handheld objects) and the \textit{Refinement phase} (table top). This shift affects the detection accuracy. Secondly, the number of images per objects (150) used during the \textit{Supervised training phase} is small. Note that, this is roughly half of the images used for the same purposes in previous works~\cite{maiettini2019b,maiettini2020}. This is done in order to show performance of the model under limited annotated data budgets. Both these aspects negatively affect the detection accuracy of the \textsc{Before refinement model} in the new conditions. However, after the refinement process, the accuracy increases remarkably, for all the 5 objects groups (\textbf{After ref.} column in Tab.~\ref{table:evaluation}), showing the effectiveness of the proposed approach. Notice that this process is performed with quite limited human intervention. The user, indeed, annotated not more than 5 images per group, drawing a total of 13 or less bounding boxes per group.

\subsection{Weak-supervision system analysis}
\label{experiments:ws_analysis}
In this section, we analyze the impact of the different components of the \textit{Weak-supervision system} during the \textit{Refinement phase}. To this aim, for each of the 5 objects groups used for the experiment in the previous section, we report in Tab.~\ref{table:ws}: (i) the comparison between the number of manual annotations required by the \textit{Weakly-supervised module} and the ones actually drawn by the human, both in terms of images and drawn bounding boxes (respectively, \textbf{Total AL queries} and \textbf{Manual annotation} columns in Tab.~\ref{table:ws}), (ii) the number of images used as self-supervision (\textbf{Total SSL} column in Tab.~\ref{table:ws}), (iii) the accuracy of the tracker (\textbf{Tracker} column in Tab.~\ref{table:ws}) and (iv) the accuracy of the bounding boxes used as ground-truth during the \textit{Refinement phase} (\textbf{Pseudo labels} column in Tab.~\ref{table:ws}).

Firstly, as it can be noted, the actual number of AL queries is greatly higher than the number of annotations provided by the human, for all objects groups. This shows the effectiveness of the  \textit{Weak-supervision system} in reducing the labeling effort with respect to previous work~\cite{maiettini2020}. This is achieved mainly by the contributions of the self-supervision and the \textit{Annotations tracker}. Regarding the former one, it can be noted in Tab.~\ref{table:ws} that the total number of annotation queries is significantly higher than the number of images chosen for self-supervision during the refinement process. This is mainly caused by the poor performance obtained by the \textsc{Before refinement model} on the table top scenario due to domain shift.
Nonetheless, in 3 out of 5 cases, self-supervision helps in reducing the amount of annotations required. However, as it can be observed in Tab.~\ref{table:ws}, the main contribution in the reduction of the required manual annotations is given by the introduction of the \textit{Annotations tracker}. Indeed, even if the total number of queries is between 134 and 200 (ranging between 652 and 800 of requested bounding boxes), the number of images that were actually annotated by the human is significantly smaller (5 or less), with a correspondent number of requested bounding boxes ranging from 10 to 13. This large reduction allows performing the refinement process on-line, during robot exploration, interactively providing the labels when requested.

Tab.~\ref{table:ws} column \textbf{Pseudo labels} shows that the accuracy of the labels considered for the refinement is not perfect (i.e. $<100\%$). One may argue that noise in the annotations provided by the self-supervision and the \textit{Annotations tracker} may have a negative impact on the refined model. However, except for one case (object group 3), the quality of the used annotations is overall quite high (i.e. $>88\%$). Moreover, the accuracy achieved in Sec.~\ref{experiments:evaluation} demonstrates that the annotation noise does not affect the refinement process, allowing to successfully recover the drop of performance obtained by the \textsc{Before refinement model} with a small price in terms of additional manual annotations.

\begin{table}[t]
	\centering
	\vspace{3mm}
	\caption{This table reports the accuracy of the \textsc{Before refinement model} and the \textsc{After refinement model} on unseen sequences. Refer to Sec.~\ref{experiments:generalization} for further details.}
	\begin{tabular}{|c|c|c|}
		\cline{1-3}
		\multicolumn{1}{|c|}{\textbf{\begin{tabular}[c]{@{}c@{}}\textbf{Objects}\\ \textbf{group}\end{tabular}}} & \multicolumn{1}{c|}{\textbf{\begin{tabular}[c]{@{}c@{}}Before ref.\\ (mAP(\%))\end{tabular}}} & \multicolumn{1}{c|}{\textbf{\begin{tabular}[c]{@{}c@{}}After ref.\\ (mAP(\%))\end{tabular}}}\\ \hline
		\multicolumn{1}{|c|}{\textbf{\#0}} &         36.0\%              &        75.8\%              \\ \hline
		\multicolumn{1}{|c|}{\textbf{\#1}} &         22.3\%              &        72.8\%              \\ \hline
		\multicolumn{1}{|c|}{\textbf{\#2}} &         42.1\%              &        81.0\%              \\ \hline
		\multicolumn{1}{|c|}{\textbf{\#3}} &         27.3\%              &        53.7\%              \\ \hline
		\multicolumn{1}{|c|}{\textbf{\#4}} &         47.8\%              &        66.0\%              \\ \hline
	\end{tabular}
	\label{table:generalization}
\end{table}

\subsection{Generalization capabilities evaluation}
\label{experiments:generalization}
Finally, in this section we evaluate the generalization capabilities of the proposed pipeline. To this aim, for each of the 5 objects groups, we evaluate the same models obtained in Sec.~\ref{experiments:evaluation}, namely, \textsc{Before refinement model} and \textsc{After refinement model}, on a new exploration sequence. This latter differs from the one used for the refinement in both the objects view poses and the arrangements. This is done in order to evaluate the accuracy of the refined detection models when presented the objects under different view poses and conditions. The results are shown in Tab.~\ref{table:generalization}.

As it can be observed, in all cases, the \textit{Refinement phase} performed in Sec.~\ref{experiments:evaluation} on one sequence, allows to improve the accuracy on the new unseen sequence as well. This demonstrates that the refinement process does not over-fit the data acquired during the \textit{Refinement phase}, but improves the generalization capabilities of the model. Moreover, the refinement process can be iteratively repeated by performing new exploration sequences, progressively improving the detection model. 
Examples of detected images from these new sequences, obtained by using both the \textsc{Before refinement model} and the \textsc{After refinement model}, are reported in Fig.~\ref{fig:detections}.


\section{CONCLUSIONS}
\label{sec:conclusions}

In this paper, we propose a unified application for efficiently training and updating an object detection on a humanoid robot.
Specifically, the proposed pipeline exploits different interaction modalities based on the interaction with a human teacher and a fixed set of robot exploratory behaviors, to quickly adapt an object detection system effectively reducing human labeling effort while retaining performance. The experimental evaluation on the real robot demonstrated the effectiveness of the proposed approach.

A limitation of this work is that the robot does not interact with the objects physically. We believe that the results can be further improved with the integration of more sophisticated active exploratory actions e.g. pushing, picking up and rotating objects to acquire new, richer views. Exploratory movements could also be linked to the learning process to determine optimal exploratory actions to obtain most informative views (i.e. \textit{best view} selection). The work presented in this paper allow the robot to iteratively adapt its vision system to novel tasks and scenarios, another direction of research would be to integrate continuous learning strategies, like e.g.~\cite{schmidt2020}.

\addtolength{\textheight}{-8.5cm}   




%



\bibliographystyle{IEEEtran}
\bibliography{bibliography}

\end{document}